# Crime Hotspot Prediction Using Deep Graph Convolutional Networks

Tehreem Zubair[1], Syeda Kisaa Fatima[1], Noman Ahmed[1], Asifullah Khan[1, 2, 3]*

[1]Pattern Recognition Lab, DCIS, PIEAS, Nilore, Islamabad, 45650, Pakistan
[2]Center for Mathematical Sciences, PIEAS, Nilore, Islamabad, 45650, Pakistan. PIEAS Artificial
[3]Intelligence Center (PAIC), PIEAS, Nilore, Islamabad, 45650, Pakistan

*Abstract*— **Crime hotspot prediction is critical for ensuring urban safety and effective law enforcement, it remains challenging due to complex spatial dependencies that are inherent in criminal activities. The traditional approaches use classical algorithms such as the KDE and SVM to model data distributions and decision boundaries. The methods often fail to capture these spatial relationships, treating crime events as independent and ignoring geographical interactions. To address this, we propose a novel framework based on Graph Convolutional Networks (GCNs), which explicitly model all of spatial dependencies by representing crime data as a graph. In this graph, nodes represent discrete geographic grid cells and edges capture proximity relationships. The spatial features from Chicago Crime Dataset are used in this system, a multi-layer GCN model is trained to classify crime types and predict high-risk zones. Our approach significantly outperforms traditional approaches, achieving 78% classification accuracy. Moreover, the model generates interpretable heat maps of crime hotspots, demonstrating the usefulness of graph-based learning for predictive policing and spatial criminology.**

## I. INTRODUCTION

Enhancing police strategies and implementing policies for crime prevention and control are greatly aided by crime prediction. Since it offers a data-driven method to increase public safety and improve resource utilization, the prediction of crime hotspots is a crucial part of modern law enforcement and urban governance. Geographical location has a significant impact on the increasing problem of urban crime. However, when it comes to simulate the complex geographical linkages of crime scenes, traditional methods typically fall short. For instance, Convolutional Neural Networks (CNNs) [2] show regions as rigid grid pixels and hence ignore nuanced spatial linkages, while Kernel Density Estimation (KDE) [1] calculate spatial intensity without taking interdependencies between geographically neighboring locations into account.

With the advancements in artificial intelligence and deep neural networks, the improvements have been achieved in crime prediction and analysis [20], [21]. It has been found that machine learning and deep learning techniques can efficiently handle large-scale crime data and identify hidden patterns for effective decision-making by law enforcement agencies [2]. Research methods like hotspot mapping techniques have indicated that historical crime data can be useful in defining high-risk areas by KDE has proven to be better in performance than some other methods of spatial mapping in some cases [16]. However, these methods primarily use historical data and do not have the capabilities of complex spatial and temporal dependencies.

The previous research which is conducted in the domain of hotspots prediction includes Crime Hotspot Prediction using clustering techniques and Crime Prediction using deep learning. Crime hotspots have been identified by aggregating spatially close incidents using a variety of clustering techniques. Chainey et al. [4] used Kernel Density Estimation (KDE) to forecast crime hot spot locations in London and depicted that spatial density could solely identify consistent high-risk areas. The drawback of clustering approaches is that it uses crimes as independent points and does not capture the cross-location dependencies (e.g., spillover effects between nearby locations [6], [10]).

Another technique which has been presented by Zhang et al. [5] has rasterized crime data into 2D grids and then applied Convolutional Neural Networks (CNNs) to predict hotspots in Chicago. Although CNNs use spatial hierarchies, they forget fine geographic connectivity due to the assumption of space as inflexible pixels. CNNs do not directly learn graph-structured representations of the location's relationship, which spatial GCNs fill [8], [12]. This paper presents a new method for crime prediction through the uses of Spatio-Temporal Generalized Additive Models (ST-GAMs) and their localized versions (LST-GAMs) in predicting and analyzing crime. Unlike existing "hot spot" analysis methods, which rely on existing crime locations, this method considers various data types in determining the underlying causes of crime. By using this method in predicting burglaries in Charlottesville, Virginia, this paper proves that LST-GAM is more accurate in predicting crimes and helps narrow down potential crime locations for more effective allocation of resources by law enforcement agencies [17].

In order to address those drawbacks, recent work has resorted to Graph Neural Networks (GNNs) [3], specifically Graph Convolutional Networks (GCNs) [6], [8], which have proven to be more effective in tasks with spatial structure. Such models can learn with graph-structured data, where spatial objects and relationships are naturally defined as nodes and edges. Moreover, the combination of deep learning and graph-based models in spatial-temporal models has demonstrated good performance in crime prediction. These models integrate both geographic adjacencies as well as the dynamics of time in order to understand crime in a more accurate manner and also enhance the accuracy of prediction.

In this paper, we present a purely spatial GCN based crime hotspot prediction model. Unlike traditional clustering methods and CNN-based approaches, the proposed model captures spatial dependencies more effectively. The main contributions of this work are as follows:

- Build a graph in which the nodes correspond to 2.2 km x 2.2 km geographic grid cells and the edges convey the spatial proximity (3 km distance).
- Makes multi-class predictions on the type of crimes and presents the high-risk areas on a map as heat maps to take action.
- The model of learning neighborhood-aware representations is effective to capture spatial interaction between regions with raw spatial data and actionable crime information.

## II. METHODOLOGY

### A. *Dataset Description*

For this study, an extensive dataset is needed that has spatial features of crime incidents, other information like the type of crime, the location description and time stamps for the crime. The City of Chicago Crime Dataset is used for this study. This dataset is a publicly available record of criminal incidents reported in the City of Chicago and is maintained by the Chicago Police Department. It provides detailed information on each reported crime, making it a valuable resource for urban crime analysis.

The data on this research has several significant strengths about crime analysis. It provides extremely accurate latitude and longitude data, enabling grid-based crime patterns. Moreover, it contains a variety of types of crimes, which enable one to analyze each type of offence in detail.

Each feature in the dataset includes geographical coordinates (WGS84) and description of locations (e.g., on the street, in the house, within the park), as well as precise times of events. Other attributes like year, month and day permit possible time analysis whereas administrative identifiers like beat, district and ward provide hierarchical geographical context. The data also consists of standardized FBI crime codes to be categorized consistently. The dataset has approximately 7 million records for the years 2001 and is up to date and it is updated on daily basis in terms of scale and coverage. It extends over the whole city of Chicago that occupies roughly 227 square miles.

Records with missing geographic coordinates (comprising less than 5% of the total dataset) were excluded to ensure the spatial integrity and reliability of the analysis.

TABLE I. SAMPLE RECORD

| date (Date when the incident occurred. This is sometimes the best estimate.) | primary_type (The primary description of the IUCR code.) | description (The secondary description of the IUCR code, a subcategory of the primary description.) | #latitude (The latitude of the location where the incident occurred. This location is shifted from the actual location for partial redaction but falls on the same block.) |
|---|---|---|---|
| 07/27/2016 02:00:00 | CRIM SEXUAL ASSAULT | NON-AGGRAVATED | 41.885888079 |

### B. *Data Preprocessing*

The raw crime dataset is prepared to ensure data quality, consistency, and suitability for network graph construction. There are four steps such as raw data ingestion and cleaning, geographic processing, feature engineering, and quality assurance. Each step can handle specific challenges while improving computational efficiency.

In the first step, the Raw Ingestion and Cleaning of dataset is loaded with robust error handling to manage potential inconsistencies, such as malformed records or encoding issues. In the pre processing step, columns are automatically cast to appropriate data types (e.g., categorical for crime types, Boolean for arrest status) to minimize memory usage, null value removal and the dataset is reduced from 7,842,937 records to 7,754,915 after filtering out entries with invalid coordinates, ensuring only spatially referenced crimes are retained.

After that, the next step is Geographical Processing that is to focus on crimes within the Chicago metropolitan area, a bounding box filter is applied (41.6° to 42.1° N, -87.9° to -87.5° W). This step removes outliers with implausible coordinates. Then, coordinates are discretized into a grid system (resolution ≈ 2.2 km) to enable spatial aggregation and network node creation, and a uniqueness check confirms that each grid cell corresponds to a distinct latitude-longitude pair, ensuring no spatial redundancy.

To enhance analytical utility, temporal and categorical features are derived:

- Spatial normalization: Latitude and longitude are scaled to a [0, 1] range to facilitate distance-based computations.
- Crime frequency normalization: Standardized using zscore normalization.
- Crime type consolidation: Rare crime categories (fewer than 1,000 instances) are grouped into an “OTHER” class to reduce sparsity and improve statistical robustness..

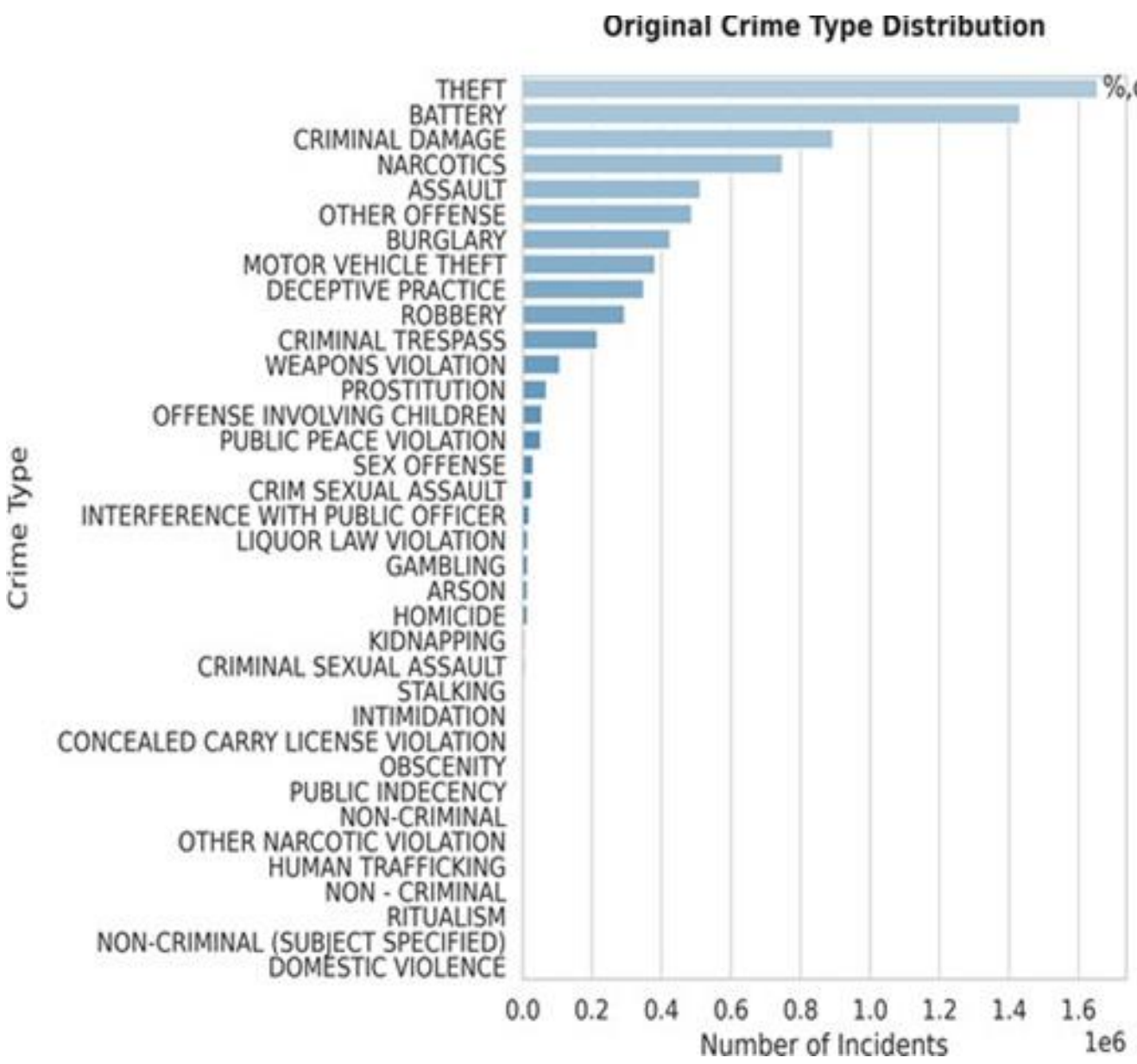

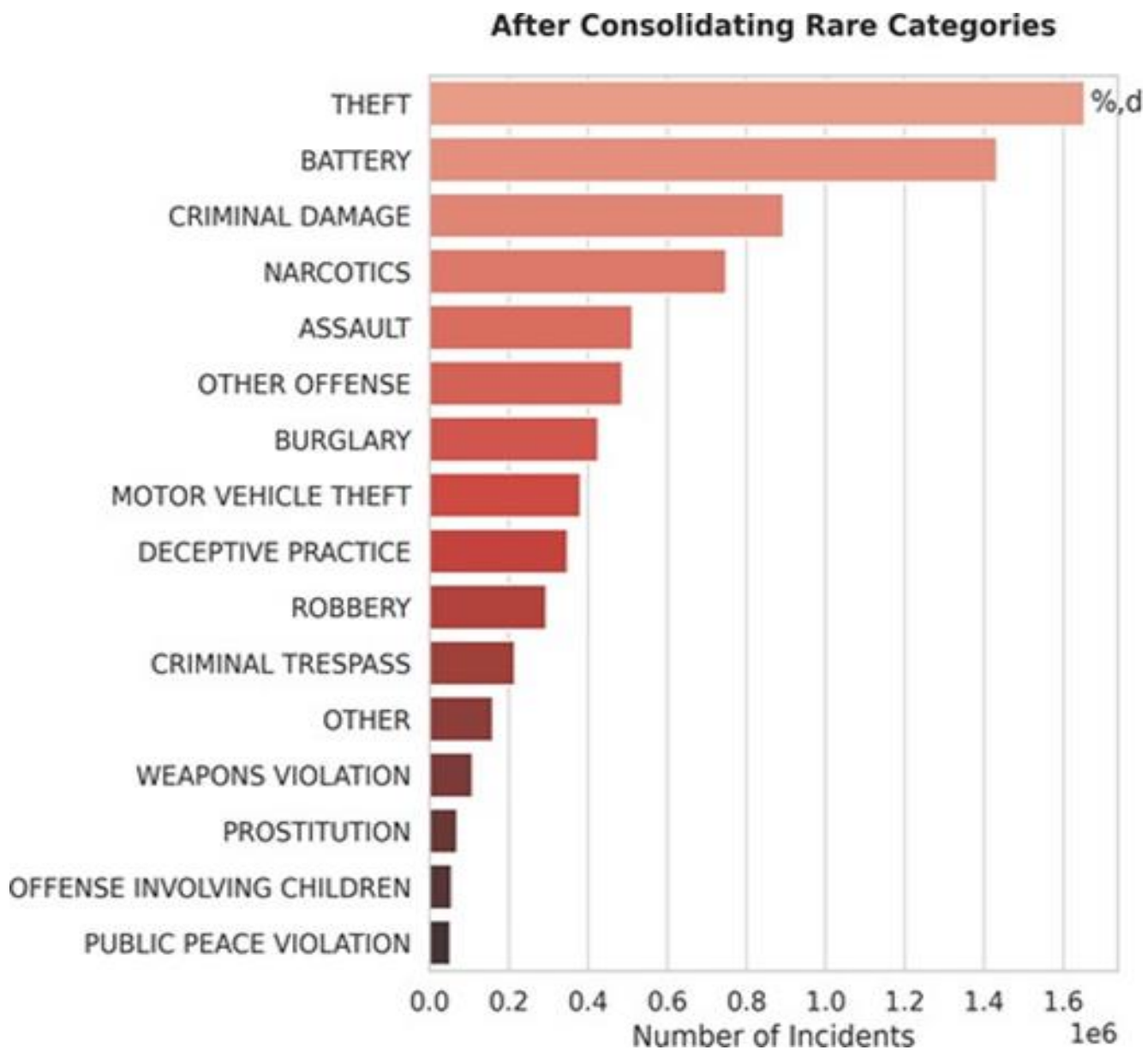

Fig. 1. Crime type distribution before (above) and after (below) consolidating 20 rare categories (<1% each) into "OTHER". Frequent crimes (THEFT, BATTERY, and CRIMINAL DAMAGE) remain distinct.

### *C. Graph Construction*

In this study, Chicago's crime patterns are represented using an undirected weighted graph G = (V, E) where:

- Nodes (V) represent 2.2 km×2.2 km grid cells covering the urban area of Chicago city. We have a total of 487 nodes in this graph.
- Edges (E) connects the previously created grids located within a proximity distance of 3 km. This distance threshold was selected based on spatial autocorrelation analysis to ensure meaningful local interactions.

(1) Node Features

- Latitude and Longitudes are normalized to a range of [0, 1]
- Crime frequency is standardized using z-score normalization

(2) Edge Weights

Computed using inverse Haversine distance between grid centroids, which is appropriate for geographic coordinates, a method validated for spatial crime networks in [10], [13]:

$$w_{ij} = \frac{1}{d_{ij}+\epsilon} \quad (1)$$

where dij is the Haversine distance between the centroids of grids i and j, and $\epsilon=10^{-6}$ is added to prevent division by zero. After creating the nodes, the next step was to connect the nodes with the help of edges.

The edge creation process was completed through a three-step pipeline which is as a follow:

(a) Raw Grid Cell Formation: As can be seen in figure 2 where an excerpt of the Edge creation process has been shown, the entire urban area was divided into 2.2 x 2.2 km grid cells.

(b) KD-Tree Based Spatial Indexing: In the second step, we need to identify the neighboring grids within a proximity distance of 3 km. To do this we have used KD-Tree Spatial Indexing. Using this method the computational complexity is reduced from O ($n^2$) to O (n log n). (Refer to Figure 2, center panel)

(c) Proximity Based Edge Creation: Edges were established between grid pairs within the threshold distance, with corresponding inverse distance weights assigned. (Refer to Figure 3, right panel)

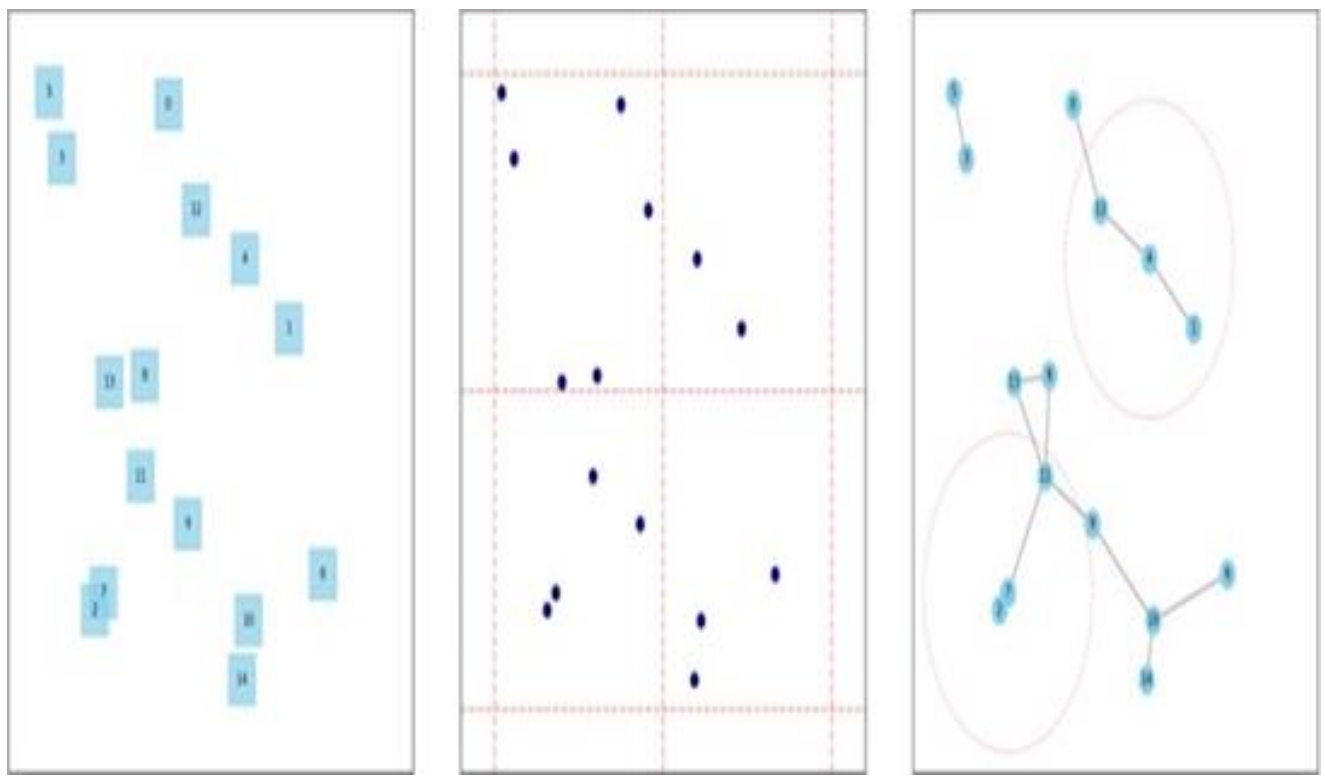

Fig. 2. Edge creation workflow showing(left) Discretization into grid cells, (middle) Spatial indexing via KD-tree, and (right) Edge formation based on proximity threshold

The statistics of the graph that are formed as a result of the defined process are given in Table 2.

TABLE II. GRAPH STATISTICS

| Hyper parameter | Value | Rationale |
|---|---|---|
| Optimizer | Adam | Adaptive Momentum |
| Learning Rate | 0.01 | Balanced convergence speed/stability |
| Weight Decay | 5e-4 | L2 Regularization |
| Epochs | 500 | Early Stopping Patience = 50 |
| Batch Size | Full Graph | Graph Level Task |
| Loss Function | Cross Entropy | Multi-class classification |

An excerpt of the graph created shows in figure 3:

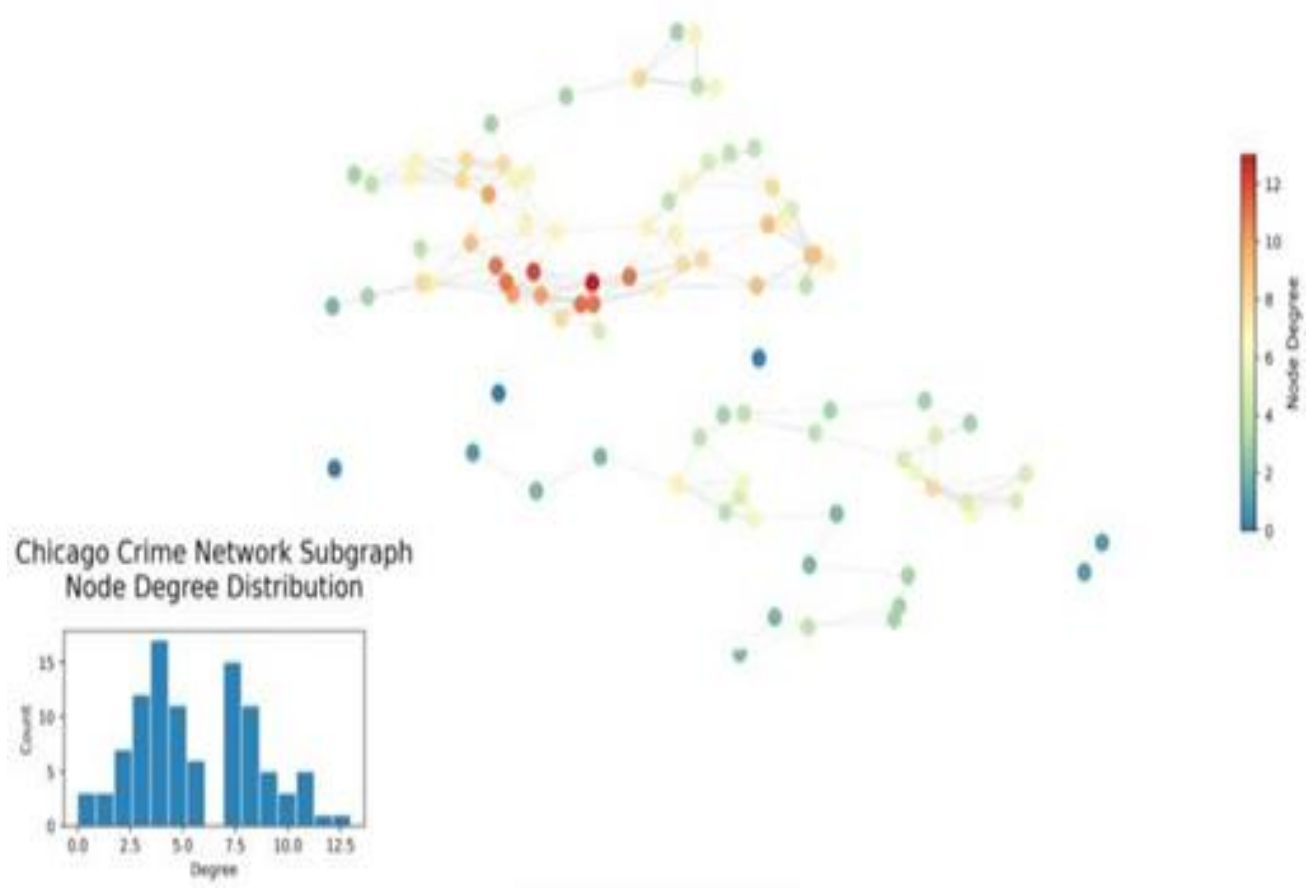

Fig. 3. An excerpt of the graph

*D. Modle Training*

After the graph was constructed, the next part is to feed the graph to GCN architecture. For Model Development we have selected two-layer model architecture. The first layer consists of 128 hidden units and is intended to represent localized spatial crime patterns, which is suitable to urban-scale graphs. The second layer combines the representations of the city at higher levels and generates the 32 classes of output that represent various types of crimes. The use of shallow architecture is a careful decision because deeper architecture can increase the possibility of over-smoothing and eliminate the capacity to capture local hotspots. Also, due to the relatively small size of the graph, namely about 500 nodes, shallow propagation can represent useful spatial relations.

The hidden dimensions of our current models are 128. The hidden dimension of 128 was chosen due to empirical assessment. The experiments with small and large dimensions of the hidden were also done. These 64-unit configurations caused underfitting, which was manifested by a drop of almost 0.05 in F1-score, whereas expansion to 256 units caused overfitting, which was evident by a 12% increase in validation loss. A dropout of 0.5 was also used to enhance generalization further and this offered effective regularization to the sparse crime data.

TABLE III. SELECTED HYPER PARAMETERS

| Metric | Value | Implication |
|---|---|---|
| Nodes | 487 | Complete coverage of urban Chicago |
| Edges | 5,832 | Average degree ≈ 11.97 |
| Density | 2.46% | Sparse spatial connections |
| Average Clustering | 0.18 | Moderate level of local connectivity |

A complete ablation study was conducted to evaluate different configurations and find the optimal one. The results are given in Table IV.

TABLE IV. PERFORMANCE COMPARISON WITH BASELINE MODELS

| Config | Hidden Dim | Layers | Drop out | Learning Rate | **F1 Score** | Train Time (min) | Params |
|---|---|---|---|---|---|---|---|
| 1 | 64 | 2 | 0.3 | 0.01 | **0.782** | 3.2 | 0.12 M |
| 2 | 64 | 2 | 0.5 | 0.01 | **0.791** | 3.2 | 0.12 M |
| **3** | **128** | **2** | **0.5** | **0.01** | **0.823** | **3.8** | **0.25 M** |
| 4 | 128 | 3 | 0.5 | 0.01 | **0.801** | 5.1 | 5.1 M |
| 5 | 256 | 2 | 0.5 | 0.001 | **0.809** | 4.9 | 0.98 M |

Out of the above five configurations here are some of the key observations: the configurations number with lesser number of hidden dimensions have shown lower performance as the F1 Score is 0.782 and 0.791 which is lower than others. Configuration 3 has shown optimal results with an F1 Score of 0.823 which is highest among all. As hidden units increased F1 Score decreased significantly. Also, the training time has increased due to abrupt increase in the number of parameters.

## III. RESULT AND DISCUSSION

The snippet of Model Performance curves shows the loss and accuracy curves for train and test sets over epochs of 600.

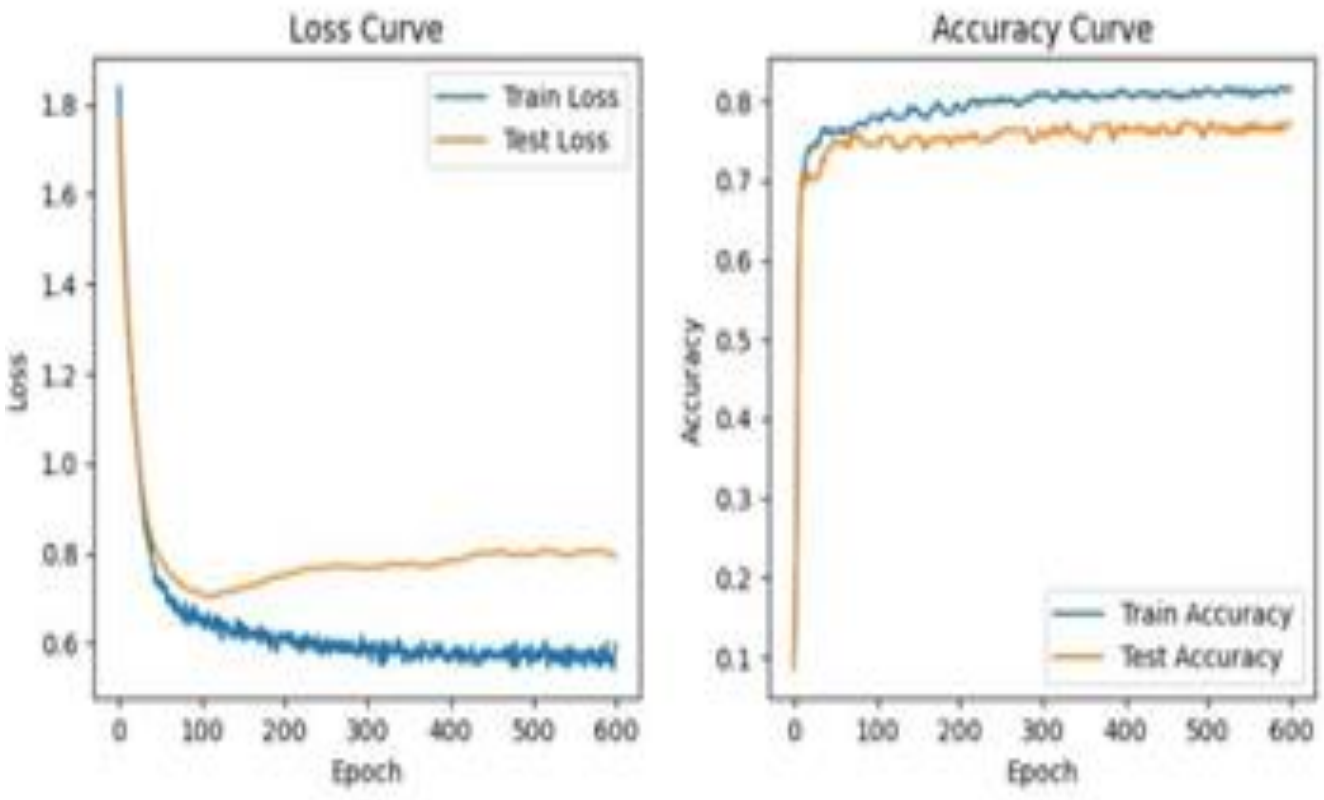

Fig. 4. Performance Curves

The training process of our two-layer GCN model was monitored using accuracy and loss curves over 600 epochs along with early stopping where patience = 50. The model has achieved convergence successfully after 300 epochs. The key observations of the training process are given below:

- Training loss has decreased steadily, and it stabilizes at around a loss value of 0.55.
- Test Loss has decreased initially but then starts to increase after 100 epochs which may indicate overfitting.
- Train Accuracy (blue) increases and stabilizes above 0.8.
- Test Accuracy (orange) increases initially then flattens around 0.72.

Due to this overfitting, the model's ability to generalize to unseen data, including other cities, may be limited.

The model then calculates crime hotspot heat maps by aggregating predicted probabilities over grid cells. In the end, high risk zones (red) are in the historical crime zones (South Side and West Garfield Park) while low risk zones (blue) are in the safer neighborhoods (Lincoln Park).

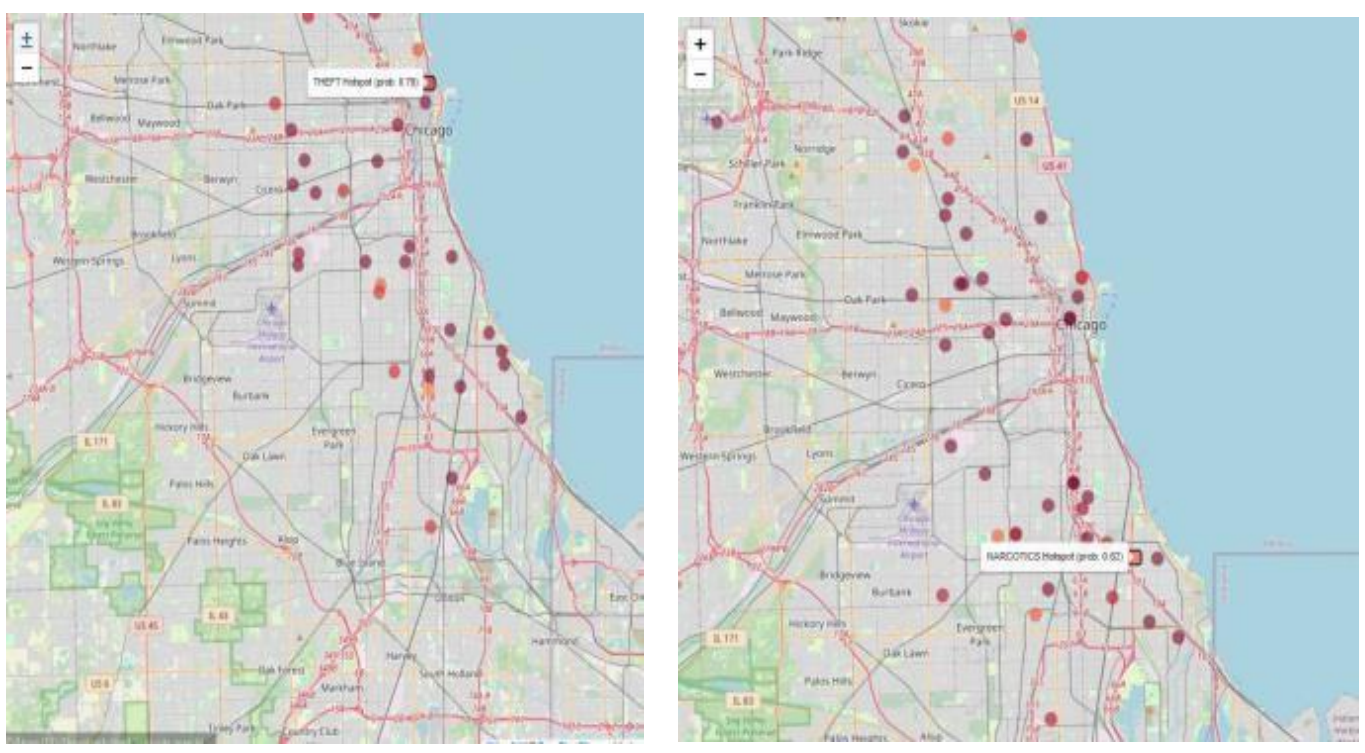

Fig. 5. Heat map for crime type "Theft (Left)" and "Narcotics (Right)"

The proposed model is compared with Kernel Density Estimation (KDE) and Support Vector Machines (SVM). We compared these two techniques with the same spatial grid discretization (2.2 km × 2.2 km) and Chicago crime dataset (2001 to present).

KDE is a nonparametric method for the detection of crime areas known as hotspots which have an estimate of the probability density function of the crime events [16]. The KDE model achieves an accuracy of 70% in binary classification, with an F1-score of 0.61, and requires approximately 12 minutes for city-wide prediction.

Support Vector Machines (SVMs) with spatial features are used as a machine learning baseline for crime prediction [17-18]. In this study, an SVM model with a radial basis function (RBF) kernel is implemented using input features that include grid-based crime frequencies and normalized spatial coordinates. To ensure a fair comparison, the same training, validation, and test split as used in the GCN model is applied. The SVM model achieves accuracy 72% with an F1-score of 0.68 and requires approximately 45 minutes for training.

The SVM technique has a number of drawbacks despite its efficacy. It depends on manually created spatial features, which might not adequately represent intricate spatial relationships.

Moreover, it cannot learn adaptive neighborhood relationships between regions and functions within a limited spatial scope. However, KDE has a number of drawbacks. Fine-grained spatial patterns are lost as a result of its tendency to over-smooth the data. Furthermore, it is incapable of adjusting to changing trends in crime over time. Furthermore, KDE ignores spatial interdependencies between nearby regions and treats each grid cell individually.

TABLE V. PERFOMANCE COMPARISON BASELINE MODELS

| Metric | GCN | SVM | KDE |
|---|---|---|---|
| Accuracy | 78% | 72% | 70% |
| F1-Score | 0.83 | 0.68 | 0.61 |
| Training Time | 4.9 min | 45 min | 12 min |
| Spatial Awareness | Explicit | Implicit | None |
| Neighborhood Effects | Learned | Ignored | Ignored |

## IV. CONCLUSION

In this study, a Spatial Deep Graph Convolutional Network (GCN) framework is presented to predict crime hotspots, which can learn the underlying spatial dependencies in urban spaces effectively. Chicago Crime Dataset is used, and our model has achieved a remarkable 78% classification accuracy on a multi- class crime prediction problem. This is a 10-16 percent increase over the baseline strategies of the traditional methods that had F1-scores of 0.61 and 0.68 respectively whereas our model achieved an F1-score of 0.83. Also, our model was faster to train and had better spatial generalization, which exemplifies its effectiveness and scale. The results clearly show that graph-based techniques, which preserve local context and identify adaptable spatial effects, have a significant advantage in spatial crime modeling. This degree of effectiveness makes our GCN model a practical, scalable, and data-driven instrument for urban safety analysis and crime prediction. To improve robustness and generalization, future work will concentrate on implementing more robust regularization techniques and testing the model on cross-city datasets.

## ACKNOWLEDGMENT

The authors would like to thank Pattern Recognition lab at DCIS, PIEAS AI Center (PAIC), and Center for Mathematical Sciences (CMS) PIEAS for providing support and computational facilities.